# Chinese–Japanese Unsupervised Neural Machine Translation Using Sub-character Level Information

**Longtu Zhang**
Tokyo Metropolitan University
6-6 Asahigaoka, Hino,
Tokyo 191-0065, Japan
zhang-longtu@ed.tmu.ac.jp

**Mamoru Komachi**
Tokyo Metropolitan University
6-6 Asahigaoka, Hino,
Tokyo 191-0065, Japan
komachi@tmu.ac.jp

## Abstract

Unsupervised neural machine translation (UNMT) requires only monolingual data of similar language pairs during training and can produce bi-directional translation models with relatively good performance on alphabetic languages (Lample et al., 2018). However, no research has been done to logographic language pairs. This study focuses on Chinese–Japanese UNMT trained by data containing sub-character (ideograph or stroke) level information which is decomposed from character level data. BLEU scores of both character and sub-character level systems were compared against each other and the results showed that despite the effectiveness of UNMT on character level data, sub-character level data could further enhance the performance, in which the stroke level system outperformed the ideograph level system.

## 1 Introduction

Although supervised Neural Machine Translation (NMT) achieved great success in recent years (Wu et al., 2016; Vaswani et al., 2017), the fact that it may fail without large parallel training data becomes a practical problem (Koehn and Knowles, 2017; Isabelle et al., 2017), especially for low-resource domains and language pairs. Lample et al. (2018) proposed an unsupervised neural machine translation (UNMT) method which only required monolingual training data to train bi-directional translation models on similar language pairs, since it heavily relied on the shared information between source and target data. They experimented on alphabetic language pairs (English–French and English–German) and showed the effectiveness of such method, that although the BLEU score is not as high as state-of-the-art supervised models, the translation quality is highly acceptable.

| Language | Word |
|---|---|
| JA-character | 風 景 |
| JA-ideograph | ⿵几虫 ⿱日京 |
| JA-stroke | ⿵⿰丿乙⿱丿⿻⿱⿰丨𠃍一⿺⿱丨一丶 ⿱⿳⿵⿰丨𠃍⿱一一 ... |
| ZH-character | 风 景 |
| ZH-ideograph | ⿵几㐅 ⿱日京 |
| ZH-stroke | ⿵⿰丿乙⿻丿丶 ⿱⿳⿵⿰丨𠃍⿱一一 ... |
| EN | landscape |

Table 1: Examples of decomposition of a Japanese word “風景” and Chinese word “风景”, both meaning “landscape” in English.

Chinese and Japanese are also similar language pairs using Chinese characters in their logographic writing systems, where there are no natural word boundaries and the characters are formed compositionally by sub-character level units, such as ideographs and strokes. Table 1 shows examples of how words in Chinese and Japanese are decomposed. Note that the ideograph and stroke sequences have higher proportion of shared parts than words, which are very useful for byte pair encoding (BPE) algorithm and shared vocabulary in machine translation systems. Having noticing this significant difference, it is worth asking whether similar NLP methods which are successful in alphabetic languages will also work for logographic languages.

The idea to integrate sub-character level information in NLP tasks is not entirely new. For example, it helps to train better word embeddings (Shi et al., 2015; Peng et al., 2017) and text classification systems (Toyama et al., 2017). Until recently, Zhang and Komachi (2018) have demonstrated that sub-character level information will help Chinese–Japanese supervised NMT systems on both encoder and decoder side. However, there is still no study on logographic UNMT systems.

**Algorithm 1:** Sharing Rate Sampling

**Data:** source/target sentences
**Input:** $r, k, N$
**Output:** source/target sentences with $r$ sharing rate ($sample$)
**Init**: $current_r, vocab, shared_vocab, sample$;
**while** $len(sample) < N$ **do**
  $current_sample \sim$ randomly sample $8 \times k$ sentences;
  calculate sentence level sharing rate $s_r$ based on $shared_vocab$;
  sort $sample$ in descending order of $s_r$;
  **if** $current_r < r$ **then**
    select top $k$ sentences;
  **else**
    select bottom $k$ sentences;
  **end**
  add selected sentences to $sample$;
  update $current_r, vocab, shared_vocab$;
  remove $current_sample$ from datasets;
**end**

Therefore, this study will try to answer the following questions:

1. Is UNMT effective for logographic language pairs such as Chinese–Japanese, especially when sub-character level information is used?
2. What is the influence of shared token rate on UNMT?

## 2 Background

### 2.1 Chinese Characters

Chinese and Japanese use structuralized strokes to form ideographs and then form characters (Japanese also has Kanas that function as phonetic letters). According to UNICODE 10.0 standard, there are 36 strokes (“一”, “丨”, “丿”, “㇏”, etc.) composing hundreds of ideographs [1], and further composing 90,000+ of different characters.

### 2.2 The UNMT Architecture

Three key principles underpin the approach to unsupervised neural machine translation (UNMT) systems.

**Shared BPE Embeddings** Instead of mapping two monolingual embeddings together (Artetxe et al., 2018), the shared BPE embedding are directly trained on concatenated source and target monolingual data. This was found efficient and effective for UNMT (Lample et al., 2018).

**Encoder–Decoder Language Models** Similar to denoising auto-encoders, the so-called encoder–decoder language models first encode monolingual data to latent representations and then decode it to the same language. The weights of the deeper layers of the encoders are often shared to enhance performance. Alternatively, an MLP discriminator can be added to discriminate the latent representations produced by different encoders.

**Back-Translation Models** UNMT borrowed the idea from Sennrich et al. (2016) and trains the back-translation models jointly in both translation directions. Specifically, for one direction, the forward NMT model first generates synthetic target data, and then translates it back to source language using the backward model.

## 3 Chinese–Japanese Sub-character Level UNMT

In addition to validate the effectiveness of UNMT in Chinese–Japanese language pair, this study will further enhances the shared information by decomposing characters into ideographs and strokes [2].

### 3.1 Character Decomposition

Both Chinese and Japanese data are encoded using UNICODE, where similar CJK characters are merged into one type. CHISE project [3] provides decomposed mapping information from CJK characters to pre-defined ideographs sequences. There are 394 ideographs and 19 special symbols for “unclear” ideographs. Besides, there are 11 additional “Ideographic Description Characters” (IDCs) to describe structural relationship between ideographs, which can help reduce the ambiguity of the decomposed data.

Based on this, we developed a decomposition tool called “textprep” to decompose character level tokenized data to sub-character level ideograph and stroke data without any ambiguity [4], which means both Chinese and Japanese data can be decomposed to ideograph and stroke sequences and compose back to character sequences. To do this, a special duplication marker (“≇”) is added if there are minor ambiguous cases. All the ideographs were manually transcribed to stroke sequences as well. Corpus without any structural information were also created for comparison reasons by removing IDCs and adding necessary duplication markers. Table 1 contains examples of different levels of character decomposition in the training corpus.

[1] The number depends on the definition of ideographs (usually around 500+).

[2] In the character level corpus we use, the average word length of Chinese and Japanese from dictionary-based tokenizers are 1.7 and 2.2, respectively, which is too short for BPE algorithm to get better shared information. Longer decomposed sequences would be preferred.

[3] http://www.chise.org/

[4] URL removed for review.

| Granularity | | JA–ZH | ZH–JA |
|---|---|---|---|
| Character | | 24.18 *(29.60)* | 29.79 *(40.00)* |
| Ideograph | w/ IDCs | 25.76* | 32.61* |
| | w/o IDCs | 25.14* *(32.00)* | 32.17* *(42.60)* |
| Stroke | w/ IDCs | **26.39*** | **32.99*** |
| | w/o IDCs | 24.75* *(32.10)* | 30.59* *(42.20)* |

Table 2: BLEU scores ($*$ for statistically significant score against baseline at $p < 0.0001$) of UNMT and supervised NMT systems (Zhang and Komachi, 2018).

### 3.2 Controlling Shared Tokens

Lample et al. (2018) have successfully made 95% of the BPE tokens in English–German language pair shared across the training set, indicating that the more proportion of token sharing, the better the UNMT systems will perform. This study sampled from same dataset with controlled token sharing rate to get a better understanding of this notion. Algorithm 1 takes controlled token sharing rate $r$, top-k value $k$ and sample size $N$ as parameters.

## 4 Experiments

To answer the research questions, two lines of experiments were performed. The Japanese–Chinese portion of Asian Scientific Paper Excerpt Corpus (ASPEC-JC (Nakazawa et al., 2016)) was used. Note that although this is a parallel corpus, we shuffled it and used it monolingually. Official training/development/testing split contains totally 670,000 Chinese and Japanese sentences for training and 2,000+ sentences for evaluating and testing. Word level BLEU are used as the metric.

**Sub-charcter level UNMT** The baseline is an UNMT system trained on Chinese–Japanese monolingual data, which are first pre-tokenized into words, and then BPE'ed using fastBPE [5]. We call it character level baseline. The experiments are to compare it against UNMT systems trained on sub-character level data, which are directly decomposed from character-level data and then BPE'ed using fastBPE. In sub-character level data, presence of structural information were also controlled by adding or removing IDCs.

[5] `https://github.com/glample/fastBPE`

| r | JA–ZH | ZH–JA |
|---|---|---|
| 0.5 | 19.72 | 25.23 |
| 0.7 | 23.60 | 28.32 |
| 0.9 | 23.04 | 28.84 |

Table 3: BLEU scores of different token sharing rate.

**UNMT with different token sharing** We sampled data ($N = 300,000$) from same monolingual corpus using Algorithm 1 with controlled token sharing rate $r$ of 0.5, 0.7 and 0.9, respectively. This is because UNMT systems trained on stroke level data with IDCs achieved the best performance in preliminary experiments.

For pre-tokenization of the data: Jieba[6] was applied to Chinese using the default dictionary; MeCab[7] was applied to Japanese using the IPA dictionary. For BPE training, the vocabulary size was set to 30,000. We uses 4-layer standard Transformer (Vaswani et al., 2017) units as our two encoders and decoders. The embedding size was 512; the hidden size of the fully connected network was 2048; the weights of the last 3 layers of the encoders are shared; the number of multi-attention head was 8. During training, the dropout rate was set to 0.1 and both vocabularies and embeddings were shared. 10% of input and output sentence were randomly blanked out to add noise to language model training. We use Adam optimizer with learning rate of 0.0001.

## 5 Results

### 5.1 Sub-character Level UNMT

Table 2 shows the results for sub-character level UNMT in both translation directions. Comparing with character-level baseline, all sub-character level models have better BLEU scores. In both stroke and ideograph systems, IDCs in the data can further enhance the performance. However, for ideograph systems, removing structural information did not decrease the performance much, comparing to a significant drop in stroke system without structural information. The best UNMT system is trained on stroke data with structural information, on both translation directions. This is in contrast with Zhang and Komachi (2018) on supervised NMT systems that when both source and target data have same granularity, ideograph systems outperformed stroke systems in both translation directions.

[6] https://github.com/fxsjy/jieba

[7] http://taku910.github.io/mecab/

| Type | Sentence |
|---|---|
| Reference–JA | 図 3 に 「 会 」 が 固有 表現 で ある か 否 か を 判定 する 2 つ の 例文 を 示し た . |
| Reference–ZH | 图 3 所示 的 是 2 个 关于 判断 “ 会 ” 是否是 固有 表达 的 例句 。 |
| Charcater–JA | 図 3 に 示す よう な 2 つ の 判断 について 「 会 」 が 固有 表現 で ある か どう か を 判断 する 例文 を 示す . |
| Character–ZH | 图 3 中 显示 了 判定 “ 会 ” 是 固有 名词 还是 有 2 个 例句 。 |
| Ideograph–JA | 図 3 に 示す よう に 2 つ の 判断 「 会 」 が 固有 表現 で ある か どう か について の 例文 を 示す . |
| Ideograph–ZH | 图 3 中 显示 了 判定 “ 会 ” 是否是 固有 名词 的 2 个 例句 。 |
| Stroke–JA | 図 3 に 示す の は , 2 つ の 判断 について 「 会 」 が 固有 表現 の 例文 で ある か どう か で ある |
| Stroke–ZH | 图 3 中 显示 了 判定 “ 会 ” 是否是 固有 表达 的 2 个 例句 。 |
| English | Figure 3 showed 2 example sentences of judging whether “会” is an inherent expression. |

Table 4: Translation examples from 3 unsupervised NMT models in 6 translation directions.

### 5.2 UNMT with Different Token Sharing

Table 3 shows the results for UNMT systems using data of different token sharing rate. When $r = 0.5$, the system recorded the lowest performance; however, when $r$ increased to $0.7$ and $0.9$, the performance differences become minor. In contrast with Lample et al. (2018), in our previous sub-character experiments, only 66% to 68% of the tokens are shared and can get relatively good BLEU score.

## 6 Discussion

This study confirms the effectiveness of UNMT systems on Chinese–Japanese smaller datasets, with greatly lower token sharing rate than Lample et al. (2018). Although the BLEU score is not as high as most RNN- and Transformer-based supervised NMT systems, it is still promising not only because of its translation quality, but also because it greatly broadens the scenario of machine translation applications.

### 6.1 Translation Quality

In both translation directions, there are a lot of synonymous expressions produced which lowered the BLEU score. However, according to native speakers' judgement, they tend to be good translations in terms of grammaticality, fluency and naturalness. For example in Table 4, the character-level system's Chinese translation “中 显示” was very close to the reference “所示” semantically, and was consistent in ideograph- and stroke-level models. A similar example would be “判断” in reference and “判定” in hypothesis. This might be due to the encoder–decoder language models, which successfully grasp the language features and express it in the translation. Consequently, if semantic-based metrics could be introduced, the performance of unsupervised NMT might be better reflected.

### 6.2 Shared Information and Proportion of Shared Tokens

Zhang and Komachi (2018) showed that shared information brought by sub-character level information can help supervised NMT systems; this study found similar phenomenon, although with different granularity preference. This is largely due to better shared information. For example in Table 4, despite the fact that translations produced by ideograph and stroke models were better than that of character model, stroke model was even slightly better than ideograph model because it translated Japanese “表現” into Chinese “表达”, which was more precise than ideograph model's “名词”. However, current unsupervised models still perform poorly on distant language pairs. If the shared information between distant languages can be improved, UNMT may work for more general purpose. Additionally, low proportion of shared tokens can harm the performance, but high proportion does not linearly improve the performance either.

## 7 Conclusion

The effectiveness of UNMT models on logographic language pair, Chinese–Japanese, is quite promising, even if using smaller training dataset. However, to evaluate its performance more accurately, better semantic-based metrics are required. Lastly, relatively high proportion of shared tokens is required for good UNMT (around 70%), but higher shared token rate seems not necessary.